\documentclass{article}
\usepackage{bbm}
\usepackage{amsmath,amssymb,amsfonts}
\usepackage{spconf,amsmath,graphicx}
\usepackage{booktabs}
\usepackage[T1]{fontenc}
\usepackage{algorithm}
\usepackage{algorithmic}
\usepackage{dsfont}
\usepackage{marvosym}
\usepackage{float}
\usepackage{graphicx}
\usepackage{hyperref}
\usepackage{url}

\usepackage{threeparttable}

\usepackage{multicol}
\usepackage{multirow}
\usepackage[capitalize]{cleveref}
\usepackage{color} 

\usepackage{verbatim}

\title{RK-core: An established methodology for exploring the hierarchical structure within datasets}
\name{Yao Lu \qquad Yutian Huang \qquad  Jiaqi Nie \qquad Zuohui Chen \qquad Qi Xuan\textsuperscript{\Letter}}

\address{Institute of Cyberspace Security, Zhejiang University of Technology, Hangzhou, 310023. China}

\begin{document}
%
\maketitle
\begin{abstract}
Recently, the field of machine learning has undergone a transition from model-centric to data-centric. The advancements in diverse learning tasks have been propelled by the accumulation of more extensive datasets, subsequently facilitating the training of larger models on these datasets. However, these datasets remain relatively under-explored. To this end, we introduce a pioneering approach known as RK-core, to empower gaining a deeper understanding of the intricate hierarchical structure within datasets. Across several benchmark datasets, we find that samples with low coreness values appear less representative of their respective categories, and conversely, those with high coreness values exhibit greater representativeness. Correspondingly, samples with high coreness values make a more substantial contribution to the performance in comparison to those with low coreness values. Building upon this, we further employ RK-core to analyze the hierarchical structure of samples with different coreset selection methods. Remarkably, we find that a high-quality coreset should exhibit hierarchical diversity instead of solely opting for representative samples. The code is available at \url{https://github.com/yaolu-zjut/Kcore}.
\end{abstract}


\section{Introduction}
Deep learning has gained tremendous achievements in a wide range of fields~\cite{brown2020language,ramesh2021zero,radford2021learning} over the past decade. At the heart of these achievements lies a common thread: the indispensability of large-scale datasets. Gaining insights into the dataset empowers researchers to unlock the full potential of data-driven AI. Nevertheless, most existing research mainly focuses on crafting advanced algorithms and building powerful models, with relatively scant attention dedicated to dataset analysis. 

Previous works has made progress in exploring dataset bias~\cite{tommasi2017deeper,fabbrizzi2022survey,wang2022revise} and analyzing dataset error~\cite{vasudevan2022does,DBLP:conf/nips/NorthcuttAM21}. For example, Wang et al.~\cite{wang2022revise} introduce a tool known as REVISE to assist in investigating a given dataset, surfacing potential object-based, person-based and geography-based biases. Northcutt et al.~\cite{DBLP:conf/nips/NorthcuttAM21} identify pervasive label errors in test sets of the most commonly-used datasets. Vasudevan et al.~\cite{vasudevan2022does} manually review and categorize every remaining error made by a few top models and provide insights into the long-tail of errors on ImageNet~\cite{deng2009imagenet}. These studies are insightful, but fundamentally limited. Because they only focus on the quality and reliability of datasets used in machine learning, while ignoring the underlying hierarchical structure within datasets. A window into the hierarchy of datasets can provide more information about the organization of datasets and the relationships among samples for better analysis and modeling. To this end, we focus on exploring the intricate hierarchical structure within datasets. 

K-core decomposition is a well-established method to study network (graph) hierarchy~\cite{alvarez2005large,kong2019k}, with applications spanning biology, ecology, computer sciences, and social sciences. As a result, our primary task is to model the dataset as a network to facilitate the application of K-core decomposition. We build upon the previous study~\cite{lu2022understanding} modeling the underlying relationship among samples in datasets as a graph. 
However, the traditional K-core decomposition suffers from monotonicity, as it assigns identical coreness to many nodes, losing a fine-grained ranking per node. To solve this problem, we introduce a novel method called RK-core. This method enables fine-grained ranking for each node by integrating round-specific information from node neighborhoods. Finally, we apply RK-core to analyze the hierarchical structure within datasets.

To summarize, we clearly emphasize our contribution as follows: 
\begin{itemize}
\item[$\bullet$] We introduce a novel approach called RK-core, designed to unravel the intricate hierarchical structure within datasets. We observe that samples with high coreness values are more representative of their respective categories than those with low coreness values. Further experiments prove that samples with high coreness values also contribute more to overall performance compared to samples with low coreness values.
\item[$\bullet$] Furthermore, we employ RK-core as an analytical tool to characterize the samples selected by different coreset selection methods. Differing from the previous conclusion, we find that a high-quality coreset should emphasize hierarchical diversity rather than exclusively focusing on representative samples.

\end{itemize}

\begin{figure}[htbp]
  \centering
   \includegraphics[width=0.99\linewidth]{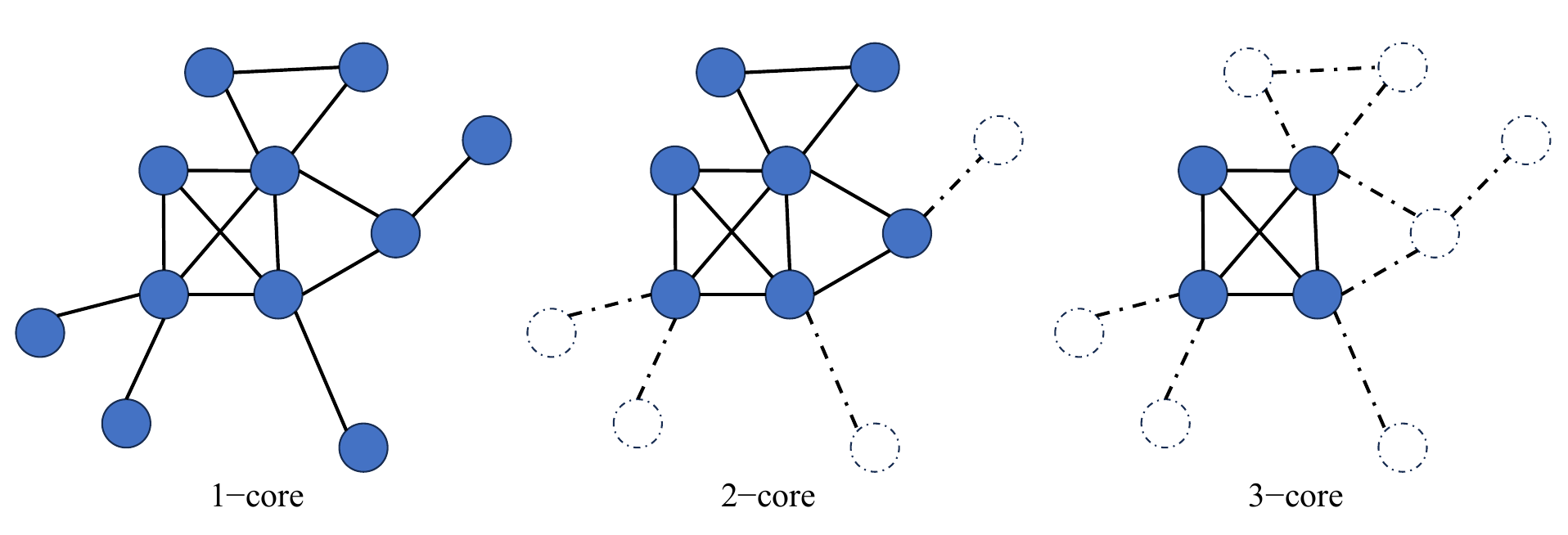}
  \setlength{\abovecaptionskip}{-2cm}
   \caption{The process of K-core decomposition. A dashed line signifies the removal of the node.}
  \label{first} 
  \vspace{-0.5cm}  
\end{figure}

\section{METHODOLOGY}
\subsection{Preliminary}
Before delving into the proposed method, it's essential to introduce the concept of K-core decomposition. In graphs, K-cores are maximal subgraphs in which each node has at least k connections to other nodes. This could be obtained by repeatedly deleting all nodes with degrees below k, as depicted in \cref{first}. The coreness of a node signifies the highest value of k for which the node belongs to a K-core. 

\subsection{Graph Construction}
In this subsection, we will elaborate on the graph construction process. Specifically, given a batch of samples $\mathcal{X} = \{x_1, x_2, \cdots, x_N\}$ sharing the same label $y_i$, we feed them into a pre-trained model $F=f_{e} \circ f_{c}$ to obtain intermediate representations $r = f_{e}(\mathcal{X}) \in \mathbb{R}^{N \times \mathit{C} \times \mathit{W} \times \mathit{H}}$. Herein, $f_{e}$ and $f_{c}$ denote the feature extractor and the classification layer, while $\circ$ corresponds to function composition. $\mathit{C}$, $\mathit{W}$ and $\mathit{H}$ denote the channel, width, and height of $r$, respectively. Subsequently, we perform dimensionality reduction on $r$ by computing the mean value for each row of $r$ in the width and height dimensions. 
To capture the underlying relationships among samples, we define a graph $\mathcal{G}=(\mathcal{V},\mathcal{E})$. This graph comprises a node set $\mathcal{V} = \{v_1, v_2, \cdots, v_N\}$ and an edge set $\mathcal{E}\subseteq\{(v_n, v_m)|v_n, v_m\in \mathcal{V}\}$, where each node in $\mathcal{V}$ corresponds to a sample and the weight of the edge ($a_{n,m}$) between nodes $(v_n, v_m)$ is calculated by the cosine similarity between their respective representations, computed as follows:
\begin{equation}
    a_{n,m} = \frac{r_n\cdot r_m^T}{||r_n||_2\cdot||r_m||_2}.
\end{equation}

Afterward, if $a_{n,m}$ surpasses the predetermined threshold $\epsilon$, we assign it the value of 1; otherwise, it is set to 0. Finally, we can obtain a unweighted and undirected sparse graph $\mathcal{G}^{\prime}$.

\subsection{RK-core}
Once we have obtained the graph, our objective is to analyze its structural properties. However, the traditional K-core decomposition has a drawback of monotonicity, where it assigns identical coreness to many nodes~\cite{carmi2007model,garas2012k,montresor2011distributed}, leading to the loss of per-node fine-grained ranking. To address the aforementioned challenge, we propose a novel variant of K-core decomposition, termed RK-core, that takes into account the round-specific information from node neighborhoods. 

\begin{algorithm}[t]
    \caption{RK-core}
    \label{alg:algorithm1}
    \textbf{Input}: $\mathcal{G}=(\mathcal{V},\mathcal{E})$: A given graph; \\
    $\mathcal{D}$: the list of degrees for all nodes in $\mathcal{V}$; \\
    $\mathcal{N}(v)$: the neighbors of nodes $v$ \\
    \textbf{Output}: $Coreness$ and $RD$ of each node in $\mathcal{V}$ \\
    \vspace{-2mm}
    \begin{algorithmic}[1] 
    \STATE Initialize $K$=1; 
    \STATE Initialize $round$=1;
    \STATE Initialize $\mathcal{V}_{loop}$= $\mathcal{V}$;
    \WHILE{$\mathcal{V}_{loop}$ is not empty}
    \STATE $\mathcal{V}_{K}=\left \{v \in \mathcal{V}_{loop}\mid \mathcal{D}(v)\le K \right \}$ 
    \FOR{each $v\in \mathcal{V}_{K}$}
    \STATE $Coreness(v)=K$;$R(v)=round$;
    \FOR{each $w\in \mathcal{N}(v)$}
    \STATE Decrease $\mathcal{D}(w)$ by 1;
    \STATE Update the corresponding degree value in $\mathcal{D}$
    \ENDFOR
    \STATE Remove $v$ from $\mathcal{V}_{loop}$ and $\mathcal{D}$
    \ENDFOR
    \STATE $round$=$round$+1
    \IF{$Min(\mathcal{D})\ge (K+1)$}
    \STATE $K$=$Min(\mathcal{D})$
    \ENDIF
    \ENDWHILE
    \FOR{each $v\in \mathcal{V}$}  \label{l1:loop_begin}
    \STATE $RD(v)$=$R(v)$
    \FOR{each $w\in \mathcal{N}(v)$}
    \STATE $RD(v)$=$RD(v)$+$R(w)$
    \ENDFOR
    \ENDFOR \label{l1:loop_end}
\end{algorithmic}
\label{alg:rd}
\end{algorithm}

Specifically, we utilize Onion Decomposition~\cite{hebert2016multi} to provide a monotonic Ranking (ODR) for nodes with the same coreness value. We argue that the ranking of a node is not only related to itself but also to its neighbors. Therefore, we define the $RD$ value as the sum of its ODR and the ODR of its neighbors. 

The overall process of RK-core is outlined in \cref{alg:rd}. Let $R(\cdot)$ be the onion decomposition ranking function, where $R(v)$ records the pruning round of $v$, i.e., ODR, $\mathcal{N}(v)$ is the neighbors of $v$ and $\mathcal{D}(v)$ is the degree of node $v$. We begin with initializing $K$=1, $round$=1 and creating an exact copy of $\mathcal{V}$, denoted as $\mathcal{V}_{loop}$, to record coreness and ODR for nodes. We select a subset $\mathcal{V}_K$ from $\mathcal{V}_{loop}$, where the degree of all nodes in $\mathcal{V}_K$ is less than or equal to $K$. Then we record the coreness and ODR for nodes in $\mathcal{V}_K$ and traverse the neighbors of each node to update their degrees, as node $v$ is removed from $\mathcal{V}_{loop}$ after finishing the traversal of its neighbors. Subsequently, $K$ is updated as the minimum degree in $\mathcal{D}$. We repeat this process until there are no nodes in $\mathcal{V}_{loop}$. Finally, we calculate the $RD$ value for each node in $\mathcal{V}$ at line \ref{l1:loop_begin}-\ref{l1:loop_end}.

\section{EXPERIMENTS}
In this section, we illustrate the feasibility of RK-core for characterizing intricate hierarchical structure within datasets. Specifically, we implement experiments on CIFAR10~\cite{krizhevsky2009learning}, CIFAR100~\cite{krizhevsky2009learning} and ImageNet~\cite{deng2009imagenet}, utilizing a pre-trained ResNet18~\cite{he2016deep} on the corresponding dataset as the feature extractor. 

In terms of model training, we use Stochastic Gradient Descent (SGD) as the optimizer with a batch size of 128, an initial learning rate of 0.1, and a training epoch of 150. Additionally, we apply learning rate decay by a factor of 0.1 at both epoch 50 and 100. To mitigate the impact of randomness, we repeat each experiment 5 times and report the mean and variance.

\begin{figure}[t]
  \centering
   \includegraphics[width=0.99\linewidth]{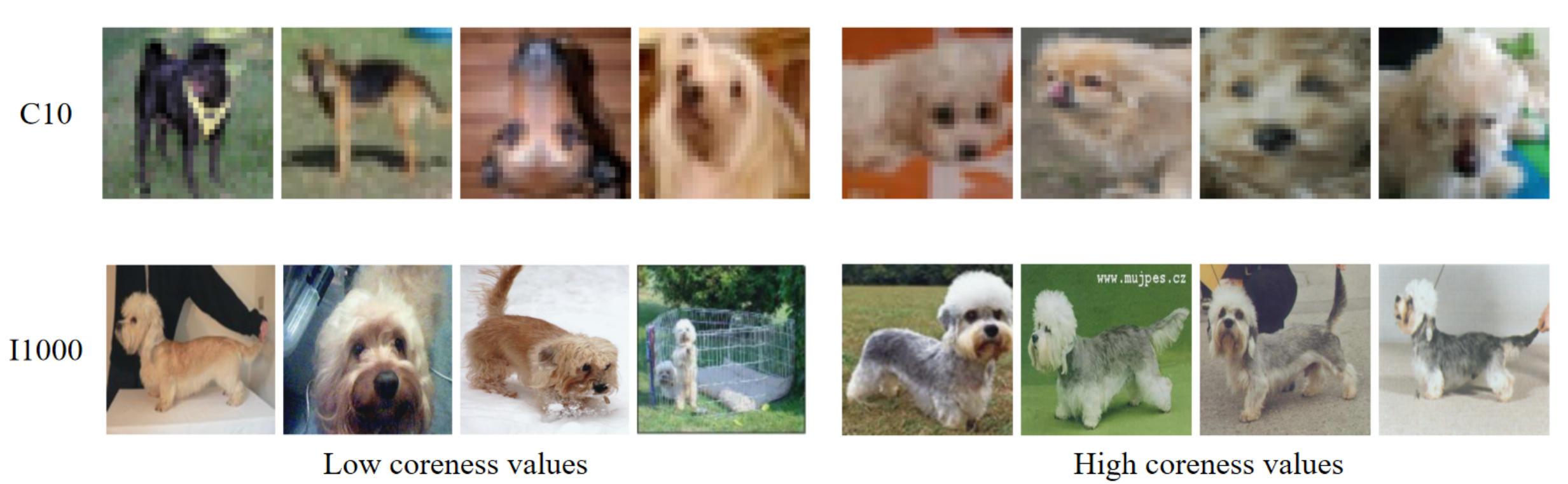}
  \setlength{\abovecaptionskip}{-2cm}
   \caption{Images corresponding to different k values. C10 and I1000 denote CIFAR10 ($32 \times 32$) and ImageNet ($224 \times 224$), respectively. For the sake of visualization, we have unified their size.}
  \label{vis} 
  \vspace{-0.5cm}  
\end{figure}

\begin{figure*}[htbp]
  \centering
   \includegraphics[width=0.9\linewidth]{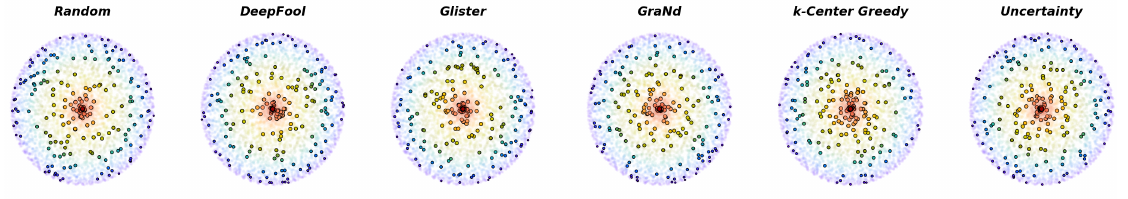}
   \caption{The visualization of graph structure using coreness values. Darker points represent samples selected by the corresponding coreset selection method. Light colored points represent samples from the original training set. The closer to the circle's center, the higher the coreness. The node size reflects the degree of the corresponding node.}
  \label{graph vis} 
\end{figure*}

\begin{figure*}[htbp]
  \centering
   \includegraphics[width=0.9\linewidth]{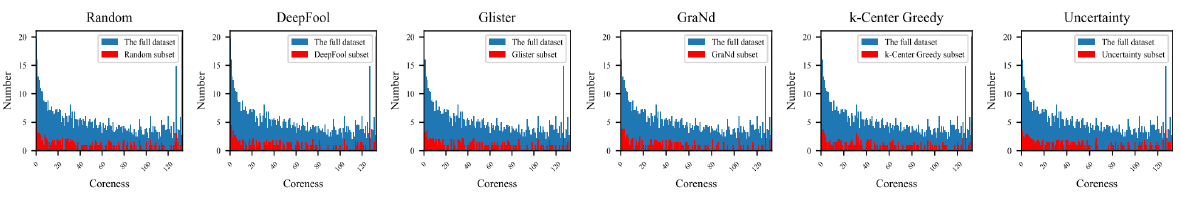}
   \caption{The distribution of coreness values. Blue and red represent the distribution of the original "airplane" class and the subset obtained by the corresponding method on CIFAR10, respectively. We plot the ordinate with a square root scale applied.}
  \label{bar} 
\end{figure*}

\begin{table}[t]
  \centering
  \caption{Performance of subsets with different coreness values. Bold entries are best results.}
  \vspace{1mm}
    \begin{tabular}{cccc}
    \toprule
    Dataset & High  & Medium & Low \\
    \midrule
    CIFAR10 & \textbf{65.1±0.1} & 61.8±0.1 & 46.2±0.2 \\
    \midrule
    CIFAR100 & \textbf{23.9±0.1} & 21.3±0.1 & 13.8±0.1 \\
    \bottomrule
    \end{tabular}%
  \label{tab:2}%
\end{table}%

\subsection{Hierarchical structure within datasets}
\label{sec:Hierarchical structure within datasets}
To explore the hierarchy in datasets, we conduct experiments on both CIFAR10 and ImageNet. Specifically, we select the "Dog" class from CIFAR10 and the "Dandie Dinmont" class from ImageNet as examples for constructing graphs. Subsequently, we apply RK-core to these graphs to obtain the coreness and the precise ranking $RD$ of each sample. We present visualizations of samples with different coreness values in \cref{vis}. We observe that samples with high coreness values exhibit greater similarity, implying a high degree of correlation in the feature semantics present within these samples. In other words, these samples tend to be more representative of their respective categories. By contrast, samples with low coreness values appear less representative and have noticeable differences in feature semantics. 

Based on these findings, a question arises: \textit{Is the contribution of samples with low coreness values the same as that of samples with high coreness values?} To answer this question, we conduct further experiments. Specifically, we select 1000 samples (20\%) with high, medium and low coreness values per class from CIFAR10, respectively. It's worth noting that if the samples have the same coreness value, we proceed to select samples according to their $RD$ values. Then we create three subsets of the original training set using these samples. According to the coreness values of selected samples, we categorize these subsets as "High", "Medium", and "Low". Finally, we employ these subsets to train ResNet18 models. As indicated in \cref{tab:2}, we find that as the coreness value increases, the performance of the corresponding subset increases synchronously. Specifically, the subset with high coreness values achieves a performance of $65.1 \pm 0.1$, whereas the subset with low coreness values only reaches $46.2 \pm 0.2$. 

To further verify the generalizability of this finding, we perform the same supplementary experiment on CIFAR100.
We select 100 samples with high, medium and low coreness values from each class on CIFAR100 and repeat the above experiment. Likewise, we can conclude that samples with high coreness values contribute more to overall performance compared to samples with low coreness values. This conclusion is consistent with our earlier explanation that samples with high coreness values tend to be more representative of their respective categories.

Overall, the qualitative and quantitative experimental results presented above provide the consistent evidence for the existence of a hierarchical structure within datasets and the effectiveness of RK-core.


\section{Use cases of RK-core}


\begin{table}[t]
  \centering
  \caption{Coreset selection performances on CIFAR10. Bold entries are best results.}
  \vspace{1mm}
  \resizebox{0.45\textwidth}{!}{%
    \begin{tabular}{cccccccc}
    \toprule
    \multicolumn{2}{c}{} & DeepFool & Uncertainty & k-Center Greedy  & GraNd & Glister & Random \\
    \midrule
    \multirow{8}[18]{*}{Fraction} & 5\%   & 41.4±0.2 & 32.3±0.1 & 58.8±0.2 & 28.3±0.3 & 40.0±0.2 & \textbf{59.6±0.2} \\
\cmidrule{3-8}          & 10\%  & 60.4±0.4 & 55.2±0.4 & 71.7±0.2 & 39.6±0.3 & 45.1±0.1 & \textbf{73.2±0.1} \\
\cmidrule{3-8}          & 20\%  & 82.1±0.3 & 80.5±0.1 & \textbf{85.4±0.3} & 67.7±0.2 & 78.5±0.2 & 83.7±0.1 \\
\cmidrule{3-8}          & 30\%  & \textbf{90.9±0.1} & 88.7±0.1 & 90.8±0.2 & 87.5±0.1 & 89.5±0.1 & 87.5±0.1 \\
\cmidrule{3-8}          & 40\%  & 92.6±0.1 & \textbf{93.0±0.1} & 92.3±0.1 & 92.6±0.2 & 92.9±0.1 & 91.7±0.2 \\
\cmidrule{3-8}          & 50\%  & 93.6±0.1 & 94.1±0.1 & 93.5±0.1 & \textbf{94.5±0.1} & 93.2±0.1 & 91.7±0.1 \\
\cmidrule{3-8}          & 60\%  & 94.8±0.1 & 94.3±0.1 & 94.1±0.1 & \textbf{94.9±0.1} & 94.6±0.1 & 92.4±0.3 \\
\cmidrule{3-8}          & 90\%  & 95.1±0.1 & 95.2±0.1 & 95.1±0.1 & \textbf{95.5±0.1} & 95.3±0.1 & 93.1±0.1 \\
\cmidrule{3-8}          & 100\% & \multicolumn{6}{c}{95.6±0.1} \\
    \bottomrule
    \end{tabular}}
    \label{tab:addlabel}
\vspace{-5mm}
\end{table}%

In the previous section, we apply RK-core to the analysis of the hierarchical structure within datasets. Herein, we expand the application of RK-core to assess the samples selected by various coreset selection methods. 

Coreset selection is a promising technique aimed at reducing computational costs. It involves selecting a small subset of the most informative training samples from a large training dataset. Models trained on this coreset are supposed to achieve similar performance as those trained on the full training set.
To this end, we first select several coreset selection methods, including Deepfool~\cite{Ducoffe2018AdversarialAL}, Uncertainty~\cite{Coleman2019SelectionVP}, k-Center Greedy~\cite{Sener2017AGA}, GraNd~\cite{Paul2021DeepLO} and Glister~\cite{Killamsetty2020GLISTERGB}. Subsequently, we reproduce these methods with the help of Deepcore\footnote{\url{https://github.com/PatrickZH/DeepCore}}~\cite{guo2022deepcore}, allowing for a fair comparison under the same experimental configuration. For each method, we select subsets in a class-balanced manner with fractions of 5\%, 10\%, 20\%, 30\%, 40\%, 50\%, 60\%, 90\% of the entire training set, respectively. \cref{tab:addlabel} reports the detailed results of different methods on CIFAR10. We observe that when the compression ratio of dataset is high, the random selection method exhibits significant competitiveness. By contrast, coreset selection methods perform better when the compression ratio is low, and different coreset selection methods exhibit varying advantages under different compression ratios. Based on the experiments above, we pose a question: \textit{Do subsets obtained by different methods have distinct hierarchies?}

To this end, we utilize RK-core to analyze the hierarchical structure of subsets obtained by different methods. Herein, we choose the ("airplane") subsets on CIFAR10 with fraction of 5\% to conduct experiments. One can select any fraction for experimentation. Furthermore, we utilize the "airplane" class from the original training set for comparative analysis. \cref{graph vis} presents the visualization of graph structure using coreness values. We observe that coreness values of the subset formed by each method are scattered across distinct numerical segments. In contrast to selecting samples with singular coreness values, opting for a diverse range of coreness values tends to enhance performance. Specifically, under the same conditions, samples with singular coreness values reach a maximum accuracy of 65.1\% on CIFAR10 (high coreness values in \cref{tab:2}), whereas samples with diverse coreness values (k-Center Greedy, 20\% in \cref{tab:addlabel}) can achieve a maximum accuracy of 85.4\%. Building upon this finding, we believe that a high-quality coreset should exhibit hierarchical diversity instead of solely opting for representative samples. To delve into the analysis of these distributions, we count the number of samples with varying coreness values and depict them in \cref{bar}. Furthermore, we find that different coreset selection methods are roughly similar to the original dataset in terms of the distribution of coreness values. The nuanced distinctions among these distributions lead to varying performance. In general, these findings confirm the effectiveness of RK-core in dataset analysis.

\section{Conclusion}
In this paper, we present an innovative approach called RK-core, to explore the intricate hierarchical structure within datasts. By deciphering the hierarchical structure concealed within datasets, we find that compared to samples with low coreness values, those with high coreness values are more representative of their respective categories. Consequently, they also play a more significant role in the training compared to the low coreness counterparts. Additional analysis reveals that, under the same conditions, samples with diverse coreness values perform better than samples with singular coreness values, which emphasizes the significance of hierarchical diversity in datasets. We aspire for RK-core to enhance the understanding of datasets through a small step forward.


\clearpage
\bibliographystyle{IEEEbib}
\bibliography{IEEE}

\end{document}